\documentclass[11pt,a4paper]{article}
\usepackage[hyperref]{eacl2021}
\usepackage{times}
\usepackage{amsmath,stmaryrd}
\usepackage{latexsym}
\usepackage{graphicx}
\usepackage{bbding}
\usepackage{pifont}
\usepackage{wasysym}
\usepackage{amssymb}
\usepackage{textcomp}

\newcommand*{\myalign}[2]{\multicolumn{1}{#1}{#2}}
\newcommand{\shortarrow}[1]{%
  \mathrel{\text{\rotatebox[origin=c]{\numexpr#1*45}{\fixed@sra}}}
}

\usepackage{microtype}

\aclfinalcopy 

\title{An Empirical Study on Measuring the Similarity of Sentential 

Arguments with Language Model Domain Adaptation}

\author{ChaeHun Park \thanks{\hspace{0.2cm}Work done while the author was an intern at Scatterlab.} \\
  School of Computing\\
  KAIST\\
  \texttt{ddehun@nlp.kaist.ac.kr} \\\And
  Sangwoo Seo \\
  Scatterlab Inc.\\
  \texttt{sangwoo@scatterlab.co.kr} \\}

\date{}

\begin{document}
\maketitle
\begin{abstract}
Measuring the similarity between two different sentential arguments is an important task in argument mining.
However, one of the challenges in this field is that the dataset must be annotated using expertise in a variety of topics, making supervised learning with labeled data expensive.
In this paper, we investigated whether this problem could be alleviated through transfer learning.
We first adapted a pretrained language model to a domain of interest using self-supervised learning.
Then, we fine-tuned the model to a task of measuring the similarity between sentences taken from different domains.
Our approach improves a correlation with human-annotated similarity scores compared to competitive baseline models on the Argument Facet Similarity dataset in an unsupervised setting.
Moreover, we achieve comparable performance to a fully supervised baseline model by using only about 60\% of the labeled data samples.
We believe that our work suggests the possibility of a generalized argument clustering model for various argumentative topics.
\end{abstract}

\section{Introduction}
Providing diverse opinions on a controversial issue is one of the most important functions of argument mining. 
To this end, methods for grouping relevant arguments within a given topic by their similarities~\cite{misra-etal-2016-measuring,reimers-etal-2019-classification,chen-etal-2019-seeing} should be developed to prohibit redundant outcomes (\textit{argument clustering}). 
This step plays a crucial role in preventing users from being overwhelmed by the number of retrieved arguments and in clarifying the inconspicuous aspects.

However, obtaining a sufficiently large labeled dataset is usually time-consuming and expensive.
A continual annotation and training process for an unseen topic is also required to avoid performance degradation.
To address this, various domain adaptation methods~\cite{li2019transferable,das2019learning,wang2019adversarial,cao2019unsupervised} have been recently explored. 
These studies aimed to appropriately transfer the knowledge within the domain or task.
In particular, several studies found that continual pretraining of a language model (e.g., BERT~\cite{devlin-etal-2019-bert} and RoBERTa~\cite{liu2019roberta}) is effective with both unsupervised domain adaptation~\cite{ma-etal-2019-domain,rietzler2019adapt} and general supervised learning~\cite{howard2018universal,gururangan2020don}.

In this study, we attempted to alleviate the low-resource problem of an argument clustering task by leveraging the recent transfer learning strategies. 
Specifically, we fine-tuned BERT using a semantic textual similarity~(STS) task to transfer the ability to measure the similarity between two sentences. 
Concurrently, we adapted the model to sentences from domains of interest. 
These two methods can drive the model to encode the proper representation, in the aspects of both domain and task.

We evaluated our approach under various conditions including the use of the labeled target dataset and the order of training. 
Experimental results show that our approach improved correlation with human-annotated similarity scores against competitive baseline models in an unsupervised setting for the Argument Facet Similarity dataset~(AFS)~\cite{misra-etal-2016-measuring}.
The sample efficiency was also improved, in that comparable performance to a fully supervised baseline model was obtained by using only about 60\% of the labeled dataset.

\begin{figure*}
  \includegraphics[width=\textwidth]{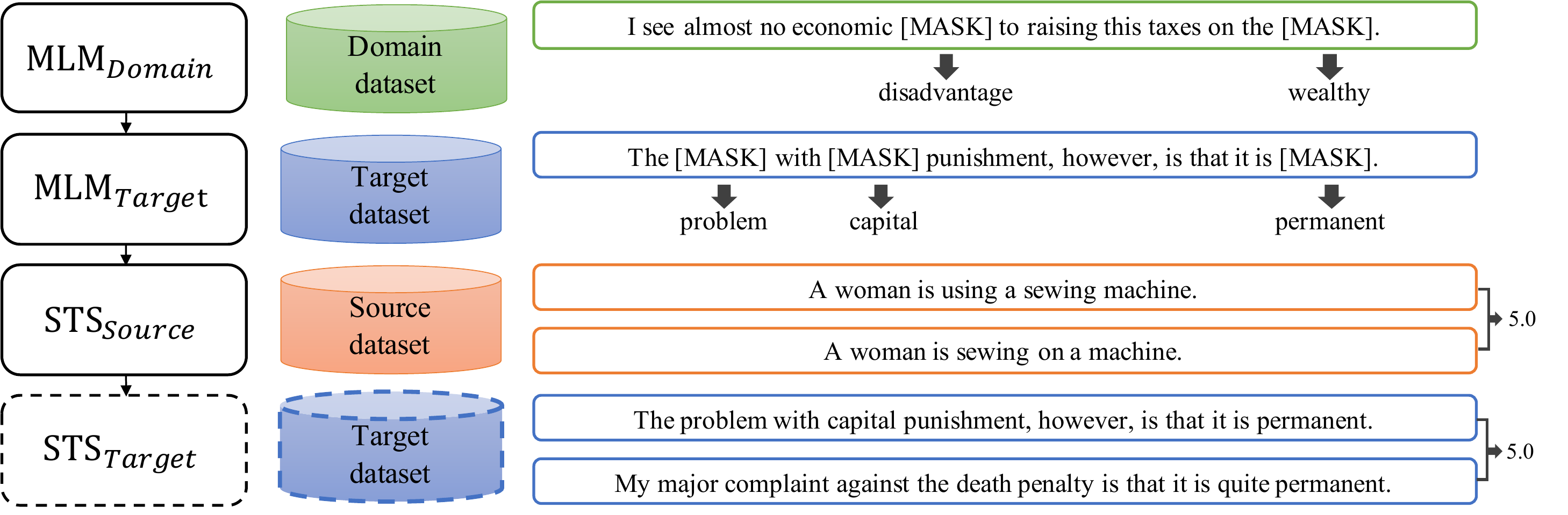}
  \caption{Overview of MLM\textsubscript{\textit{domain}}$\shortrightarrow $MLM\textsubscript{\textit{tgt}}$\shortrightarrow$STS\textsubscript{\textit{src}}. STS\textsubscript{\textit{tgt}} is only performed in a supervised setting.}
    \label{fig:pipeline}
\end{figure*}

Our contributions are as follows: (1) We formulate the task that measures the similarity between sentential arguments into an unsupervised domain adaptation problem.
(2) We investigate various strategies to adapt the pretrained language model into the desired domain and task. 
(3) Our proposed approach constantly achieves higher correlation scores than strong baseline models in unsupervised, low-resource, and fully-supervised settings. 

\section{Approach}
\label{approach}
We formulated the argument clustering task as measuring the similarity between two sentential arguments. 
For this, we used a sentence-BERT~\cite{reimers2019sentence} as our basic architecture.
When two sentences were given, each sentence was individually transformed into a fixed-size vector by a shared single BERT.
We used cosine similarity to measure the similarity score between two sentences. 

Our approach consists of two different methods~(Fig.~\ref{fig:pipeline}). The first method adapts the pretrained BERT to domains of interest through self-supervised learning~(Section~\ref{mlm-section}). 
The other method fine-tunes the sentence-BERT architecture for an STS task with a dataset other than our target dataset~(Section~\ref{sts-section}). 

\subsection{Masked Language Modeling for Domain Adaptation}
\label{mlm-section}
We used masked language modeling~(MLM) to adapt BERT to our target data distribution. 
This strategy randomly masks the tokens of an input sequence and trains the model to correctly predict the original token based on its unmasked context. 
This process was expected to shift the distribution of the model toward the desired domain and enable the model to extract the better representations of target sentences. This adapted BERT is then used to get semantically meaningful sentence embeddings.

For this step, we used two unlabeled corpora with different characteristics, following Gururangan et al.~\shortcite{gururangan2020don}.
The first corpus is composed of sentences from the target dataset itself, to adapt the model to the target distribution. We denote this adapted BERT by \textbf{MLM\textsubscript{\textit{tgt}}}. 
The second is a larger corpus that contains arguments on various topics other than ones in the target dataset. 
This domain-level adaptation conveyed more general knowledge of argumentation to the model. 
This model is denoted by \textbf{MLM\textsubscript{\textit{domain}}}.

\subsection{Transfer Learning from a Similar Task}
\label{sts-section}
We performed supervised learning for a sentence-pair similarity regression task using STSb dataset~\cite{cer-etal-2017-semeval}. 
The underlying hypothesis here was that the ability to measure the similarity between relatively common sentences could be transferred to our narrow range of domains.
This can be regarded as a typical unsupervised domain adaptation training, where only the labeled dataset from the source domain~(STSb) exists.
This model is denoted by \textbf{STS\textsubscript{\textit{src}}}.

\subsection{Training Procedure}
We considered different combinations among the abovementioned methods to find the best curriculum strategy. 
If two or more methods were used, each method was processed sequentially.
For instance, if STS\textsubscript{\textit{src}} and MLM\textsubscript{\textit{domain}} methods were chosen, two different models can be made based on the order of training (MLM\textsubscript{\textit{domain}}$\shortrightarrow$STS\textsubscript{\textit{src}} and STS\textsubscript{\textit{src}}$\shortrightarrow$MLM\textsubscript{\textit{domain}}). 
These models were either fine-tuned for the target task (if labeled data existed), or used directly for the target task. 
We did not investigate the combinations of MLM\textsubscript{\textit{domain}} following the other two methods~(STS\textsubscript{\textit{src}} and MLM\textsubscript{\textit{tgt}}) since the number of data samples available is much larger for MLM\textsubscript{\textit{domain}}~(2.3M) than for the others~(6K and 8K, respectively).

\begin{table*}
\centering
\begin{tabular}{ccccc}
\hline
\bf Name  & \bf MLM & \bf Fine-tuning & \bf Size & \bf Model\\ \hline
AFS~\cite{misra-etal-2016-measuring}  & \checkmark & $\triangle$ & 6,000~(pair) & MLM\textsubscript{\textit{tgt}}\\ 
Reddit~\cite{hua-wang-2018-neural}  & \checkmark &  & 2.3M~(sent.) & MLM\textsubscript{\textit{domain}} \\
STSb~\cite{cer-etal-2017-semeval}  &  & \checkmark & 8,628~(pair) & STS\textsubscript{\textit{src}} \\ 
\hline
\end{tabular}
\caption{Dataset details. Fine-tuning on AFS was performed in a supervised setting only.}
\label{tab:dataset}
\end{table*}
\section{Experimental Setup}
\label{experiment}
We used AFS dataset~\cite{misra-etal-2016-measuring} as our main target dataset for the argument clustering task. 
This dataset contains sentential arguments on three controversial topics (\textit{gun control}, \textit{death penalty} and \textit{gay marriage}). 
STSb dataset was used as a source domain for STS task~\cite{cer-etal-2017-semeval}. 
In AFS and STSb datasets, similarity scores are annotated on a scale from 0 to 5. 
For domain-level MLM, we used the dataset crawled from Reddit \url{r/ChangeMyView} subcommunity~\cite{hua-wang-2018-neural}~\footnote{https://www.reddit.com/r/changemyview}. 
In this community, users post their replies to change the viewpoints of other users about various controversial topics. 
The details of each dataset are described in Table~\ref{tab:dataset}.

We used Adam optimizer~\cite{kingma2014adam} with the initial learning rate set to 2e-5 and applied gradient clipping with a maximum norm of 1~\cite{pascanu2013difficulty}. 
We trained MLM on AFS for 10 epochs, as well as on Reddit for 5 epochs.
We fine-tuned STS task for 5 epochs on both STSb and AFS datasets.
In MLM, we randomly dropped 15\% of the tokens in a sentence. 
We used dropout with a rate of 0.1~\cite{srivastava2014dropout}. 
We set a random seed to 42 for every experiment.

We compared our approach with the following baseline models:
\textit{BERT}~\cite{devlin-etal-2019-bert}\footnote{The pretrained BERT (\url{bert-based-uncased}) by Huggingface~\cite{Wolf2019HuggingFacesTS} was used for our experiments.}, \textit{Glove }~\cite{pennington-etal-2014-glove}, \textit{InferSent}~\cite{conneau2017infersent}, \textit{Universal Sentence Encoder}~\cite{cer2018universal}. 
The similarity score between two sentence embeddings was measured by cosine similarity.
As previously mentioned, the original BERT and all of our methods are used as an encoder of sentence-BERT to get a sentence embedding of each sentential argument.

\begin{table}[]
    \centering
    \begin{tabular}{c c c}
        \hline \bf Model & \bf $r$ & \bf $\rho$ \\ 
        \hline \multicolumn{2}{l}{\textbf{Unsupervised - Baseline}} & \\ \hline
        \myalign{l}{GloVe} & .1443 & .1632 \\
        \myalign{l}{InferSent-GloVe} & .2741 & .2699 \\
        \myalign{l}{InferSent-FastText} & .2741 & .2699 \\
        \myalign{l}{BERT} & .3464 & .3413 \\
        \myalign{l}{Universal Sentence Encoder} & .4445 & .4358 \\
        \hline \multicolumn{2}{l}{\textbf{Unsupervised - Ours}} & \\ \hline
        \myalign{l}{MLM\textsubscript{\textit{tgt}}} & .3947 & .4071 \\
        \myalign{l}{STS\textsubscript{\textit{src}}} & .4002 & .3881 \\
        \myalign{l}{STS\textsubscript{\textit{src}}$\shortrightarrow$MLM\textsubscript{\textit{tgt}}} & .4195 & .4203 \\
        \myalign{l}{MLM\textsubscript{\textit{domain}}} & .4654 & .4564 \\
        \myalign{l}{MLM\textsubscript{\textit{tgt}}$\shortrightarrow$STS\textsubscript{\textit{src}}} & .4662 & .4454 \\
        \myalign{l}{MLM\textsubscript{\textit{domain}}$\shortrightarrow$MLM\textsubscript{\textit{tgt}}} & .4707 & .4648 \\
        \myalign{l}{MLM\textsubscript{\textit{domain}}$\shortrightarrow$STS\textsubscript{\textit{src}}} & .4767 & .4699 \\
        \myalign{l}{MLM\textsubscript{\textit{domain}}$\shortrightarrow$STS\textsubscript{\textit{src}}$\shortrightarrow$MLM\textsubscript{\textit{tgt}}} & .4779 & .4685 \\
        \myalign{l}{MLM\textsubscript{\textit{domain}}$\shortrightarrow$MLM\textsubscript{\textit{tgt}}$\shortrightarrow$STS\textsubscript{\textit{src}}} & \textbf{.5209} & \textbf{.5085} \\
        \hline
        \end{tabular}
        \captionof{table}{Evaluation results in an unsupervised setting. The highest score is highlighted in bold.}
        \label{tab:unsupervised}
\end{table}

\section{Results and Analysis}
\label{result}
We evaluated Pearson correlation ($r$) and Spearman's rank correlation coefficient~($\rho$) for each method,
following previous works~\cite{misra-etal-2016-measuring,reimers-etal-2019-classification}.
The average scores over a 10-fold cross-validation setup are reported.

\subsection{Results on Unsupervised Setting}
Table~\ref{tab:unsupervised} presents the evaluation results of each model in an unsupervised setting. Among the baseline models, \textit{Universal Sentence Encoder} showed the best performance. 
From the result of our methods, we observed that all of our proposed single models achieved better performance in both metrics than the original BERT model. 
A combination of any method followed by others performed better than single methods. 
In particular, our best model~(MLM\textsubscript{\textit{domain}}$\shortrightarrow$MLM\textsubscript{\textit{tgt}}$\shortrightarrow$STS\textsubscript{\textit{src}}) improved Pearson correlation by 50.37\% and Spearman's rank correlation by 48.98\% compared with BERT.
These results indicate that our proposed method can effectively measure the similarity of sentential arguments in the unsupervised setting. 
We also found that even if the same methods were used, performance differed significantly depending on the order of training
~(For instance, MLM\textsubscript{\textit{tgt}}$\shortrightarrow$STS\textsubscript{\textit{src}} and STS\textsubscript{\textit{src}}$\shortrightarrow$MLM\textsubscript{\textit{tgt}}). 
We speculate that this is because fine-tuning the model with a proper downstream task is required in the final process of training, which should be further investigated in future work.

\subsection{Analysis on Sample Efficiency}
\begin{figure}[t!]
\centering
\includegraphics[width=0.48\textwidth]{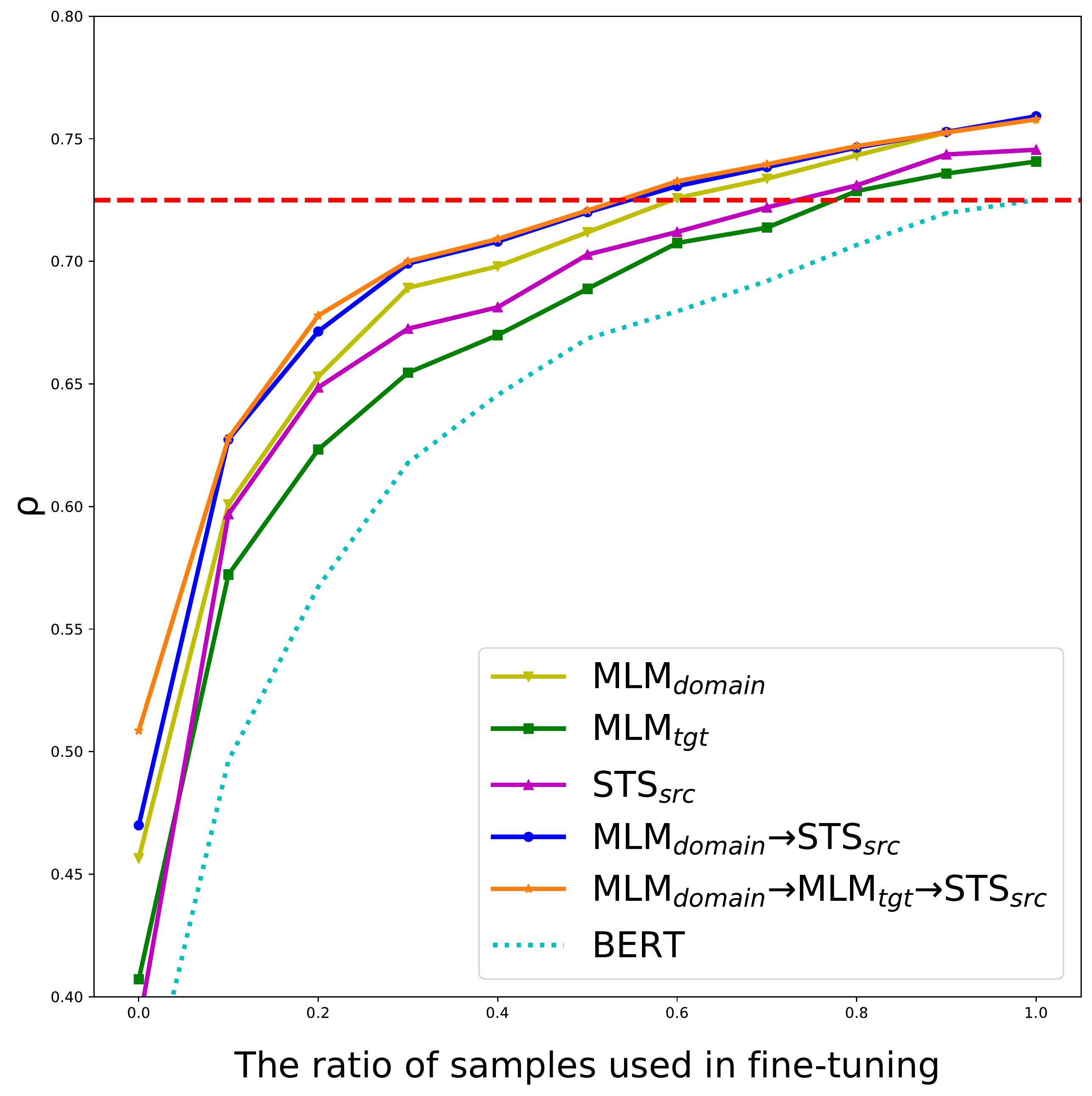}
\captionof{figure}{Spearman's rank correlation~($\rho$) for each model as a function of the ratio of data samples used in fine-tuning. The dotted red line indicates \textit{BERT} in a supervised setting.}
\label{fig:ratio}
\end{figure}

To verify the sample efficiency of the proposed methods, we further fine-tuned each model using AFS dataset by increasing the ratio of labeled data samples by 10\%. The results are depicted in Fig.~\ref{fig:ratio}. 
Our models reached the performance of the fully supervised BERT by using only about 60\% of the labeled data. 
In the fully supervised case, our best model improved both metrics by 3-4\% upon BERT~(Table~\ref{tab:supervised}).

\subsection{Verifying the Effectiveness of Our Method}
One natural question is whether the performance improvement in our approach was due to increase in the number of training samples, regardless of the training details. 
To verify this, we used the MNLI dataset~\cite{MNLI} to train BERT by either an MLM~(MLM\textsubscript{\textit{MNLI}}) or a supervised NLI classification task~(NLI\textsubscript{\textit{MNLI}}). 
The training epochs for MLM and NLI fine-tuning were set to 5 and 3, respectively. 
The results are presented in Table~\ref{tab:mnli}.
As can be observed, supervised training using the MNLI dataset slightly dropped the performance of BERT, regardless of whether the labeled AFS dataset was used.
Masked language modeling improved the performance compared to the original BERT, although not superior to any of our methods.

\begin{table}[t!]
\centering
\begin{tabular}{c c c}
\hline \bf Model & \bf $r$ & \bf $\rho$ \\ \hline
\multicolumn{2}{l}{\textbf{Supervised}} & \\ \hline
\myalign{l}{BERT} & .7520 & .7249 \\
\myalign{l}{MLM\textsubscript{\textit{tgt}}} & .7637 & .7407 \\
\myalign{l}{STS\textsubscript{\textit{src}}} & .7655 & .7455 \\
\myalign{l}{MLM\textsubscript{\textit{tgt}}$\shortrightarrow$STS\textsubscript{\textit{src}}} & .7756 & .7549 \\
\myalign{l}{MLM\textsubscript{\textit{domain}}$\shortrightarrow$STS\textsubscript{\textit{src}}} & .7776 & \textbf{.7591} \\
\myalign{l}{MLM\textsubscript{\textit{domain}}} & .7786 & .7581 \\
\myalign{l}{MLM\textsubscript{\textit{domain}}$\shortrightarrow$STS\textsubscript{\textit{src}}$\shortrightarrow$MLM\textsubscript{\textit{tgt}}} & .7789 & .7579 \\
\myalign{l}{MLM\textsubscript{\textit{domain}}$\shortrightarrow$MLM\textsubscript{\textit{tgt}}} & \textbf{.7801} & .7570 \\

\hline
\end{tabular}
\captionof{table}{Evaluation results in supervised setting. The highest score is highlighted in bold.}
\label{tab:supervised}
\end{table}
\begin{table}[t!]
\centering
\begin{tabular}{c c c}
\hline \bf Model & \bf $r$ & \bf $\rho$ \\ \hline
\multicolumn{2}{l}{\textbf{Unsupervised}} & \\ \hline
\myalign{l}{NLI\textsubscript{\textit{MNLI}}} & .3325 (-.0139) & .3030 (-.0383) \\
\myalign{l}{MLM\textsubscript{\textit{MNLI}}} & .3772 (+.0308) & .3804 (+.0391) \\ 
\hline \multicolumn{2}{l}{\textbf{Supervised}} & \\ \hline
\myalign{l}{NLI\textsubscript{\textit{MNLI}}} & .7367 (-.0153) & .7024 (-.0225) \\
\myalign{l}{MLM\textsubscript{\textit{MNLI}}} & .7593 (+.0073) & .7375 (+.0126) \\
\hline
\end{tabular}
\captionof{table}{Evaluation results for MNLI dataset. NLI\textsubscript{\textit{MNLI}} and MLM\textsubscript{\textit{MNLI}} denote the model trained by the original NLI task and MLM, respectively. The numbers in parentheses represent differences from the original BERT.}
\label{tab:mnli}
\end{table}

\section{Conclusion}
\label{conclusion}
We investigated a way of leveraging transfer learning to address the low-resource problem of the sentential argument clustering task. 
To this end, we used two simple methods to adapt the pretrained language model to the target data distribution and the task itself. 
Experimental results showed that there was a reasonable performance gain in the unsupervised setting, and also improvement in the sample efficiency in the supervised setting.
Empirical results imply that our approach could be used to train a more efficient and accurate model for argument clustering.

As future work, we intend to extend our approach to a general clustering setup, not limited by a sentence-pair similarity. We also plan to investigate if such knowledge could be transferred for other tasks as well in argument mining, for instance, stance classification~\cite{bar-haim-etal-2017-stance} and evidence detection~\cite{thorne2019fever2}.

\bibliography{eacl2021}
\bibliographystyle{acl_natbib}

\end{document}